%% file: arxiv.tex
\documentclass{article}
\usepackage[preprint]{neurips_2026}

\usepackage{afterpage}
\usepackage{amsmath}
\usepackage{amssymb}
\usepackage{booktabs}
\usepackage{bm}
\usepackage[utf8]{inputenc} 
\usepackage[T1]{fontenc}    
\usepackage{graphicx}
\usepackage{hyperref}       
\usepackage{url}            
\usepackage{booktabs}       
\usepackage{amsfonts}       
\usepackage{nicefrac}       
\usepackage{microtype}      
\usepackage{multirow}
\usepackage{paralist}
\usepackage{soul}
\usepackage{subcaption}
\usepackage[dvipsnames]{xcolor}

\definecolor{coral}{HTML}{F25F5C}
\newcommand*{\corrtext}{Corresponding author: Ahmet Onur Akman, \url{onur.akman@uj.edu.pl}}

\title{PC3D: Zero-Shot Cooperation Across Variable Rosters via Personalized Context Distillation}
\input{sections/_authors}
\begin{document}
\maketitle


\input{sections/00_abstract}

\input{sections/01_introduction}
\input{sections/02_otc}
\input{sections/03_methodology}

\input{sections/04_results}
\input{sections/05_conclusions}


\begin{ack}
This work was financed by the European Union within the Horizon Europe Framework Programme (ERC Starting Grant COeXISTENCE no. 101075838). Views and opinions expressed are however those of the authors only and do not necessarily reflect those of the European Union or the European Research Council Executive Agency. Neither the European Union nor the granting authority can be held responsible for them.
\end{ack}


\bibliographystyle{plain}
\bibliography{references} 


\newpage
\appendix
\section*{Appendix}

\input{sections/a_additional_results}

\input{sections/b_novelty}
\input{sections/c_discussion}

\input{sections/d_repro}



\end{document}

%% file: sections/_authors.tex
\author{
    Ahmet Onur Akman$^{1,2,}$\thanks{\corrtext}\hspace*{0.15cm},
    \textbf{Rafał Kucharski}$^{2}$
    \vspace{0.2cm}\\
    \textnormal{$^1$ Doctoral School of Exact and Natural Sciences, Jagiellonian University, Kraków, Poland}\\
    \textnormal{$^2$ Faculty of Mathematics and Computer Science, Jagiellonian University, Kraków, Poland}
}

%% file: sections/00_abstract.tex

\begin{abstract}
Cooperative multi-agent reinforcement learning often assumes a fixed execution team, yet many decentralized systems must operate with varying numbers of active agents during deployment. We study this setting under episodic roster variation: each episode is executed by a set of homogeneous agents, with the team size varying across episodes. Agents act only from local histories, without execution-time communication, privileged coordinators, or online retraining. Therefore, effective cooperation requires each agent to recover relevant context about the active team and adapt its behavior accordingly. To this end, we propose \textbf{PC3D} (Personalized Central Coordination Context Distillation), a method for training decentralized policies to recover and use personalized coordination context from local interaction histories. During training, a set-structured centralized teacher compresses the active team into coordination tokens and personalizes them into agent-specific contexts, which are distilled into decentralized policies. At execution, each agent predicts its own context from local history and adaptively uses it to condition decision-making. Across three cooperative MARL benchmarks, PC3D achieves higher returns than the evaluated baselines with both seen and unseen roster sizes, and ablations attribute these gains to both context distillation and adaptive context use.
\end{abstract}

%% file: sections/01_introduction.tex
\section{Introduction}
\label{sec:intro}

Multi-agent reinforcement learning (MARL) studies how multiple decision-makers learn to act in shared environments, making it a natural framework for cooperative control problems that require coordinated behavior \cite{zhang_marl, oroo_cmarl, gron_marl}. As the application domains expand, cooperative multi-agent systems are increasingly expected to operate in settings where agents cannot rely on execution-time coordination mechanisms \cite{zhang_decentralized}. Centralized training with decentralized execution (CTDE) has become the dominant framework for addressing this tension in cooperative MARL (CMARL) \cite{ctde_review}. Classical value-factorization methods such as VDN and QMIX improve decentralized control by constraining how a centralized training objective decomposes into per-agent utilities \cite{vdn,qmix}. Centralized-critic methods such as MADDPG, COMA, and MAPPO instead use training-time information to stabilize policy learning while preserving decentralized execution \cite{maddpg, coma, mappo}. 

Although CTDE methods have substantially advanced the field, they typically assume a fixed execution team. This leaves a structural gap for \emph{open-team cooperation} (OTC), where the team size may vary during deployment. We refer to the set of agents active in a given episode as a \emph{roster}, and study \emph{episodic roster variation}: each episode is executed by a fixed roster, but the roster size may change across episodes. Our setting involves homogeneous agents and assumes fully decentralized execution under partial observability, without execution-time communication, global observations, or online retraining. OTC naturally arises in many decentralized control problems, including robot teams, which can be restructured to meet operational requirements \cite{robotic_teams, robot_patrol}; warehouse systems, where the infrastructure can be rescaled depending on corporate objectives \cite{robot_warehouse}; and autonomous vehicle routing, where the fleet size may evolve with changing demand and adoption rates \cite{av_fleets_1, av_fleets_2, urb}.

Several lines of work address variation in team composition in CMARL \cite{open_env_survey} with different constraints on the execution model or the task structure. Agent–entity graph methods \cite{transferable_akshat} learn policies over agents and entities by relying on graph message passing. SOG \cite{sog} organizes agents into temporary conductor-follower groups, exchanging summarized messages during execution. COPA \cite{copa} uses a privileged coach with an ``omniscient'' view to distribute strategies during both training and execution. MIPI \cite{mipi}, building on REFIL \cite{refil}, regularizes reliance on team-related information by assuming that the designer could decompose agent states into team-related ($s^-$) and -unrelated ($s^+$) components. In contrast, we explore whether a method can address the OTC problem natively within the CTDE setting without changes to the execution model, with centralized information available only during training and execution relying solely on each agent's local history. By keeping the execution contract fixed and treating the method as the variable of interest, we ask whether methodological changes alone can improve zero-shot cooperation across roster sizes.


In this setting, the core challenge is not merely learning effective coordination for a given task, but maintaining it across different, possibly unseen roster sizes at execution time. This aspect resembles Ad Hoc Teamwork (AHT), which is an adjacent problem concerned with adapting learners to unfamiliar teammates \cite{stone_adhoc, rahman_gpl, wang_oaoht}. Although the two settings share structural challenges, AHT primarily focuses on mitigating coordination failures caused by unfamiliar teammate policies, whereas OTC requires leveraging additional teammates when new cooperative opportunities arise. This makes CTDE methods a viable option to train such policies, although their standard form does not account for changing cooperation regimes across roster sizes. To that end, this study explores \emph{whether CTDE methods can improve cooperation across varying team sizes and zero-shot generalization to unseen ones by leveraging a centralized team representation, which can provide personalized and locally recoverable coordination signals while preserving fully decentralized execution}.

Existing methods provide ingredients for this goal. \emph{Teacher-student} methods such as CTDS and PTDE show that centralized guidance can be distilled into decentralized agents \cite{zhao_ctds} and that this guidance should be \emph{agent-personalized} \cite{chen_ptde}; however, they leave open how this signal should be formed, personalized, and used under the OTC setting. On scalability and architectural compatibility with changing roster sizes, attention-based and permutation-invariant critics have shown that centralized representations can handle unordered agent collections \citep{iqbal_maac, liu_pic}, with Deep Sets and Set Transformers providing the underlying design principles \cite{zaheer_deepsets,lee_settransformer}.
However, these pieces do not, by themselves, solve OTC: a set-compatible critic improves the centralized training signal but does not automatically provide the decentralized policy with a reusable notion of coordination across roster sizes.

\begin{figure}[t]
    \centering
    \includegraphics[width=0.95\linewidth]{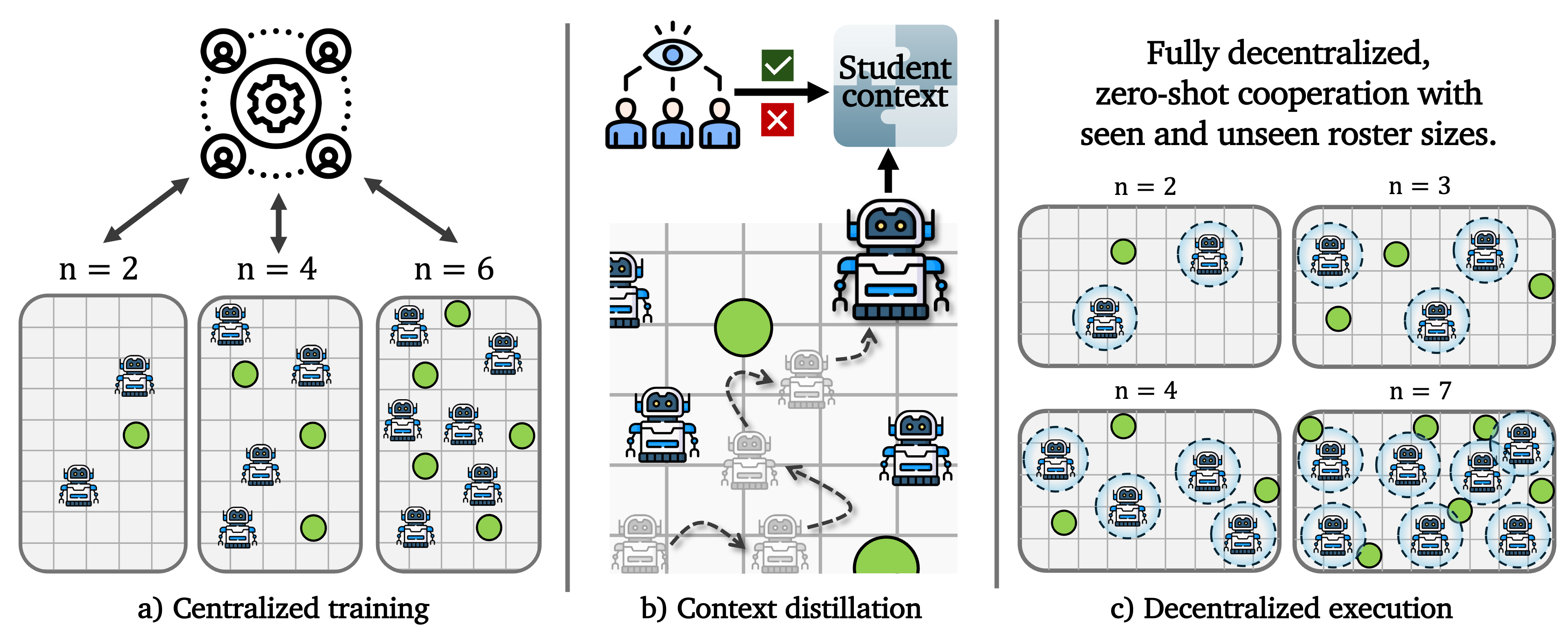}
    \caption{\textbf{PC3D at a glance.} PC3D trains with centralized information over a distribution of roster sizes for a given cooperative task (\textbf{a}). During training, the centralized teacher provides personalized coordination contexts, which decentralized agents learn to recover from local interaction histories (\textbf{b}). At execution, agents act only from local histories, without communication or retraining, and coordinate across both seen and held-out roster sizes (\textbf{c}).}
    \label{fig:intro}
\end{figure}

Building on these ideas, this paper introduces \textbf{PC3D} (\textbf{P}ersonalized \textbf{C}entral \textbf{C}oordination \textbf{C}ontext \textbf{D}istillation): a method for improving CTDE learners under episodic roster variation by (i) extracting a compact team-level coordination summary using a set-structured central module, (ii) personalizing that summary into locally recoverable contexts, and (iii) distilling it into decentralized policies that learn how and when to rely on it. We instantiate it on top of a MAPPO backbone, although the idea can be extended to other CTDE learners. During training, a centralized set critic embeds the active team as an unordered set, compresses it into a small number of \textit{coordination tokens} via token-based cross-attention, and produces personalized \textit{per-agent teacher contexts}. These teacher contexts are used for training agents to infer team context estimates from local interactions, while the coordination tokens support centralized value estimation. At execution, each agent still acts only on its local observation history while also predicting a \emph{student coordination context} to adaptively condition policy features. As illustrated in Figure \ref{fig:intro}, training PC3D over a distribution of roster sizes within the same cooperative task enables decentralized policies to recover relevant team context from local histories and coordinate under both seen and held-out roster sizes.

This research has been structured around our central hypothesis: \textbf{For a given task structure, compact team coordination representations can be personalized into locally recoverable agent contexts and distilled into decentralized executors, enabling enhanced team-context awareness at the agent level for stronger cooperation across seen rosters and better zero-shot generalization to unseen ones}. To rigorously study it, we first formalize the problem (Section \ref{sec:otc}), propose a method that reflects this methodological intent (Section \ref{sec:method}), conduct evaluations tailored to confirm our hypothesis (Section \ref{sec:main_results}), and perform further analyzes to strengthen our conclusions (Section \ref{sec:ablations} and Appendix \ref{sec:additional_results}). Our evaluations across three cooperative MARL benchmarks show that PC3D achieves the highest returns on both seen and held-out rosters, consistently improving its MAPPO backbone by a clear margin and outperforming the IPPO and PIC{\scriptsize-MAPPO} baselines. Moreover, ablations attribute these gains to the full distillation-adaptive conditioning mechanism, not merely to adding a stronger centralized critic. 

\paragraph{Contributions.}
\begin{compactitem}
    \item We provide a new formalization for the variable-roster cooperation problem (where each episode is executed by varying teams of homogeneous agents) using a family of cooperative tasks induced from a common template.
    \item We propose \emph{personalized central coordination context distillation} as a solution for open-team cooperation and instantiate it on top of a MAPPO backbone.
    \item We evaluate our method across three cooperative MARL benchmarks with varying roster sizes. We highlight the added value of our method by comparing it to three MARL baselines.
    \item We perform ablations to test the marginal gains of distillation and adaptive policy conditioning. Moreover, we offer additional insights on whether the distilled context is recoverable from local history and is meaningfully used for decision-making.
\end{compactitem}


%% file: sections/02_otc.tex
\section{Open-Team Cooperation}
\label{sec:otc}

We study fully cooperative, partially observable tasks in which a set of agents must act autonomously to optimize a shared objective. Such tasks are generally formalized as a \textit{Decentralized Partially Observable Markov Decision Process} (Dec-POMDP) \cite{decpomdp}. Although this formulation is useful for describing a task instance, it is insufficient to capture the higher-level OTC objective of learning generalizable cooperative policies in the presence of episodic roster variability. 

We focus on tasks with homogeneous agents (sharing the same action and observation spaces) and refer to a set of agents admitted in an episode as a \textbf{roster}. We represent a cooperative task with different rosters using a family of Dec-POMDPs induced from a common \textbf{environment template}. An environment template $E$ describes the shared structural properties of a space of Dec-POMDPs and can be formalized as a tuple:
\begin{equation*}
E = \left(\mathcal{S}, A, O, \mathcal{U}, \mathcal{R}, \Gamma, \gamma \right),
\end{equation*}
where $\mathcal{S}$ is the shared state-description schema of the task, $A$ and $O$ are the shared per-agent action and observation spaces, $\mathcal{U}$ is the shared cooperative objective, $\mathcal{R}$ is the set of \textit{admissible rosters}, $\Gamma$ is the \textit{roster-indexing mechanism} that instantiates roster-specific Dec-POMDPs with the task semantics defined by the template, and $\gamma$ is the discount factor. For each roster $r \in \mathcal{R}$, the template induces a roster-indexed Dec-POMDP
\begin{equation*}
    \mathcal{M}_r
    =
    \Gamma(r)
    =
    \left(r, S_r, A, O, P_r, \Omega_r, R_r, \rho_r, \gamma\right),
\end{equation*}
where $r$ is the roster (active agent set), $S_r$ is the state space induced by the template for that roster, $P_r$ is the transition kernel, $\Omega_r$ is the joint observation kernel, $R_r$ is the shared reward function, and $\rho_r$ is the initial-state distribution. Thus, $\Gamma$-induced Dec-POMDPs share the task semantics and cooperative objective of $E$, while allowing roster-dependent dynamics and observation/reward structure.

\textbf{Optimality.}
To define optimality for a given environment template $E$, we use the notion of \textbf{policy generators}. A policy generator $G$ is a mapping of the given environment template and roster pair to a joint decentralized policy ($G(E,r) = \pi_r \in \Pi_r$). This object describes the family of decentralized policies induced across rosters and not an execution-time coordinator. A policy generator $G^*$ is optimal for the given environment template $E$ if
\begin{equation*}
    G^*(E, r) \in \arg \max_{\pi_r \in \Pi_r} J_r(\pi_r), \qquad \forall r \in \mathcal{R}, 
\end{equation*}
where $J_r(\pi_r)$ is the expected discounted return of a decentralized joint policy $\pi_r$ for the roster $r$. This defines the ideal roster-wise objective. Since training separate policies across $\mathcal{R}$ is often impractical as it scales with $|\mathcal{R}|$, we study whether a shared policy mechanism trained on $\mathcal{R}_s \subset \mathcal{R}$ can approximate this objective and generalize zero-shot to held-out rosters in $\mathcal{R} \setminus \mathcal{R}_s$.

%% file: sections/03_methodology.tex
\section{PC3D: Personalized Central Coordination Context Distillation}
\label{sec:method}

We hypothesize that achieving strong generalization across different-roster task instances in a partially observable, fully decentralized setting requires enhanced context awareness and adaptive decision-making. We therefore propose \textbf{PC3D}, a CTDE extension for open-team settings that preserves the practical conveniences that make CTDE attractive: centralized information can shape learning during training, while execution remains fully decentralized with the same local observation interface.

PC3D builds on the teacher-student CTDE idea of distilling centralized signals into decentralized executors \cite{zhao_ctds, chen_ptde}, but targets a distinct structural limitation. For instance, PTDE \cite{chen_ptde} distills personalized global information into decentralized agents to improve local decision-making. While this is useful for fixed-roster cooperation, extending this idea to the OTC setting introduces additional requirements, which we tailor PC3D to explicitly address: the centralized representation should be responsive to roster variability, transferable across roster-induced cooperation regimes, and personalized in a way that remains tied to agent-observable features to support recoverability. Therefore, the method employs components to (i) produce a global representation that compactly summarizes the coordination context for the active team, (ii) from which to produce per-agent teacher contexts that include useful and recoverable coordination cues for decision-making, (iii) use context distillation to recover these contexts from local information at execution time, and (iv) adaptively condition agent policies on the estimated context. This study introduces PC3D atop a MAPPO backbone (illustrated in Figure \ref{fig:method}) with parameter-shared (to reuse across varying numbers of agents) and recurrent (using GRUs \cite{gru} to mitigate partial observability \cite{drqn, qmix}) actor networks.

\afterpage{
\begin{figure}[t]
    \centering
    \begin{minipage}[c]{0.55\linewidth}
        \centering
        \includegraphics[width=\linewidth]{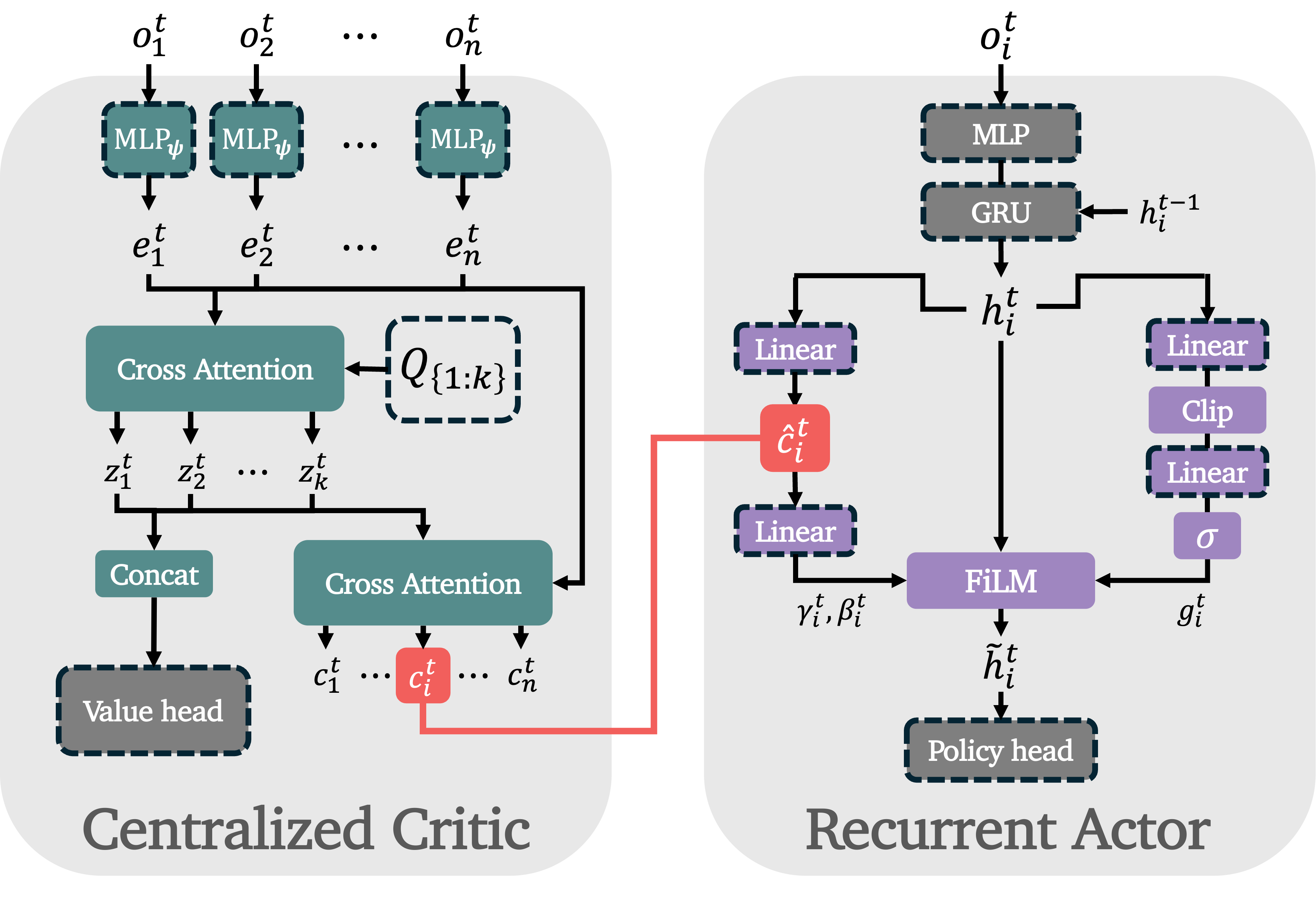}
    \end{minipage}
    \hfill
    \begin{minipage}[c]{0.43\linewidth}
        \caption{\textbf{PC3D{\footnotesize-MAPPO} architecture.} PC3D extends MAPPO with a critic/teacher module (left) and a context-conditioned actor (right). The critic encodes agent observations with a shared encoder, read by learned query tokens $Q_{1:k}$ through cross-attention to produce coordination tokens $z_{1:k}^t$, used for team value prediction. This representation is personalized in a secondary cross-attention into per-agent teacher contexts $c_i^t$. The actor uses recurrent features $h_i^t$ to predict a student context $\hat c_i^t$ to FiLM-modulate policy features, controlled by the context-reliance gate $g_i^t$. The dashed boxes indicate trainable components and the \textbf{\textcolor{coral}{coral}} connection denotes the distillation path.}
        \label{fig:method}
    \end{minipage}
\end{figure}}

\subsection{Centralized coordination context and personalization}

We replace the fixed-width centralized critic with a permutation-invariant set critic for architectural compatibility with varying team sizes. At each training step, the set critic receives the observations of the active agents and encodes them individually with a shared encoder:
\begin{equation*}
e_i^t = \phi_\psi(o_i^t), \qquad i \in r,
\end{equation*}
where $r$ is the active roster and $o_i^t$ is the observation of agent $i \in r$. Then, within the teacher module, a small number $K$ of learned query vectors ($q_k$) attend to these observation encodings through a single-head cross-attention layer with identity projections to produce $K$ coordination tokens ($z_k^t$):
\begin{equation*}
\alpha_{kj}^t
=
\operatorname{softmax}_{j \in r}
\!\left(
\frac{q_k^\top e_j^t}{\sqrt d}
\right),
\qquad
z_k^t
=
\sum_{j \in r} \alpha_{kj}^t e_j^t,
\qquad
k=1,\dots,K.
\end{equation*}
By using cross-attention with a fixed number of trainable query vectors, we enforce an information bottleneck that yields a compact coordination summary. This is intended to make the representation more transferable across rosters by biasing the critic toward team-level factors most useful for value estimation rather than overly granular roster-specific details. The coordination tokens $Z^t$ are concatenated into a fixed-width team representation and passed to the value head to predict the centralized team value. In parallel, the per-agent observation encodings attend back to the coordination tokens in a secondary cross-attention layer to produce per-agent teacher contexts ($c_i^t$), which are personalized coordination contexts used in context distillation:
\begin{equation}\label{eq:teacher_context}
\eta_{ik}^t
=
\operatorname{softmax}_{k = 1,\dots,K}
\!\left(
\frac{(e_i^t)^\top z_k^t}{\sqrt d}
\right),
\qquad
c_{i}^t
=
\sum_{k=1}^{K} \eta_{ik}^t z_k^t,
\qquad
i \in r .
\end{equation}
This construction is permutation-invariant for the value branch and permutation-equivariant for the per-agent teacher contexts, which allows for assigning one personalized context to each active agent independently of agent ordering. Implementing the teacher module within the centralized critic enables the value loss to shape the teacher's parameters to identify useful team features for value estimation. The learned query vectors ($Q$) first extract team-level factors from the set of agent observation embeddings ($E^t$), shaped by the value objective, so that the coordination tokens ($Z^t$) tend to encode compact patterns that matter at the collective level. Then, the secondary attention provides each agent with a personalized context ($c_{i}^t$) by retrieving the subset of these latent factors most aligned with the agent's embedding ($e_i^t$). Using dot-product attention with identity projections keeps this readout \emph{similarity-based}, which is a deliberate inductive bias intended to reduce the risk that the context is overly shaped by the value loss through unnecessary learnable flexibility and to make it more likely to remain structured, recoverable, and tied to agent-observable features.

\subsection{Decentralized context recovery and adaptive conditioning}

PC3D employs shared-parameter actor networks for reuse by a variable number of agents. To enable agents to recover and leverage teacher context under partial observability, we equip the actor networks with context-estimation and feature-modulation paths. First, recurrent actor features ($h_i^t$) undergo two linear transformations to produce the agent's context ($\hat c_i^t$) and context reliance control signal ($\rho_i^t$) estimates:
\begin{equation} \label{eq:student_context}
\hat c_i^t = W_c h_i^t + b_c,
\qquad
\rho_i^t
=
\mathrm{clip}\!\left(w_u^\top h_i^t + b_u,\; \rho_{\min},\; \rho_{\max}\right).
\end{equation}

We clip $\rho_i^t$ to stabilize early training and reduce premature gate ($g_i^t$ below) saturation. $\rho_i^t$ is then converted into a gating scalar to control the modulation of the recurrent features in a \textit{Feature-wise Linear Modulation} (FiLM) \cite{film} with the agent's context estimation:
\begin{equation} \label{eq:condition}
\left[ \gamma_i^t;\beta_i^t \right] = W_f \hat c_i^t + b_f,
\qquad
g_i^t = \sigma(a_g \rho_i^t + b_g),
\qquad
\tilde h_i^t
=
h_i^t \odot (1 + g_i^t \gamma_i^t) + g_i^t \beta_i^t,
\end{equation}

where $\gamma_i^t$ and $\beta_i^t$ are the scaling and shifting terms for the FiLM modulation; $a_g$ and $b_g$ are scale and offset control parameters for the context reliance gating. The resulting transformed hidden features ($\tilde h_i^t$) are then fed into the policy head.

We use feature modulation (instead of a concatenation such as $[h_i^t;\hat{c}_i^t]$) so that context estimation can adaptively influence the policy features without competing with them as a separate input stream. Moreover, the context estimate $\hat{c}_i^t$ is distilled from the teacher context $c_i^t$ (Eq. \ref{eq:distill_loss}) shaped for team-value estimation, so the way this information should affect action selection is not fixed a priori. The actor learns, through the policy objective, how (by $\gamma_i^t$ and $\beta_i^t$) and to what extent (by $g_i^t$) the recovered context should shape policy features.

\subsection{Training objective}

PC3D{\scriptsize-MAPPO} retains the standard MAPPO optimization components, including PPO-style clipped policy updates, centralized value regression, entropy regularization, and GAE-based advantage estimation \cite{ppo,gae,mappo, entropy_ppo}, and extends the learning objective with a single distillation term.

Let $\bar c_{i}^t$ denote the \textit{detached} personalized teacher context used as the distillation target for agent $i$ at time $t$ (from Eq. \ref{eq:teacher_context}), and let $\hat c_i^t$ denote the student context predicted from local history (from Eq. \ref{eq:student_context}). We train the student context with a smooth $L_1$ (Huber) distillation loss:
\begin{equation} \label{eq:distill_loss}
\mathcal{L}_{\mathrm{distill}}
=
\frac{1}{|\mathcal{D}|}
\sum_{(t,i)\in\mathcal{D}}
\ell_{\mathrm{Huber}}
\!\left(
\hat c_i^t,\;
\bar c_{i}^t
\right),
\end{equation}
where $\mathcal{D}$ is the set of valid agent-time decision pairs in the minibatch. Huber distillation loss allows the distillation to recover from large early-stage teacher-student misalignment without weakening regression in well-aligned contexts. In practice, we use the exponential moving average of the teacher to stabilize the distillation target during policy updates (controlled by $\tau$ in Table \ref{tab:hyperparams}).

Then the full PC3D{\scriptsize-MAPPO} objective becomes
\begin{equation} \label{eq:final_loss}
\mathcal{L}
=
\mathcal{L}_{\mathrm{PPO}}
+
\lambda_V \mathcal{L}_{V}
-
\lambda_H \mathcal{L}_{H}
+
\lambda_{\mathrm{distill}} \mathcal{L}_{\mathrm{distill}},
\end{equation}
where $\mathcal{L}_{\mathrm{PPO}}$, $\mathcal{L}_{V}$, and $\mathcal{L}_{H}$ are the standard MAPPO actor, value, and entropy terms, respectively.

The objective is optimized over a training distribution over a subset of admissible rosters for the same cooperative task. This exposes the learner to multiple roster sizes during training, while preserving the evaluation goal of decentralized execution on both seen and held-out rosters.

Importantly, the context-reliance estimation receives no direct supervision. It is optimized only with respect to the policy objective so that the model learns, via return maximization, how strongly and under what conditions the context estimates should influence action selection.

%% file: sections/04_results.tex
\section{Results}
\label{sec:results}

\subsection{Experimental setup}
\label{sec:setup}

\begin{figure}[ht]
    \centering
    \begin{minipage}{0.6\linewidth}
    \centering
    \begin{subfigure}[t]{0.3\linewidth}
        \centering
        \includegraphics[width=\linewidth]{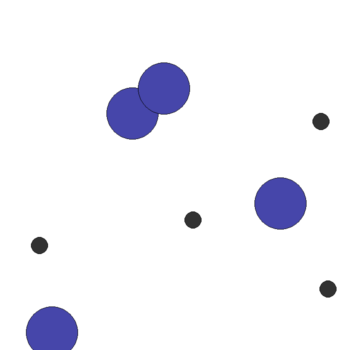}
        \caption{Spread}
        \label{fig:spread_ss}
    \end{subfigure}
    \hfill
    \begin{subfigure}[t]{0.3\linewidth}
        \centering
        \includegraphics[width=\linewidth]{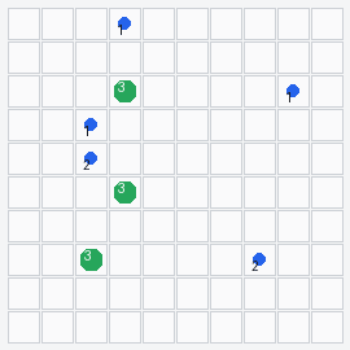}
        \caption{LBF}
        \label{fig:lbf_ss}
    \end{subfigure}
    \hfill
    \begin{subfigure}[t]{0.3\linewidth}
        \centering
        \includegraphics[width=\linewidth]{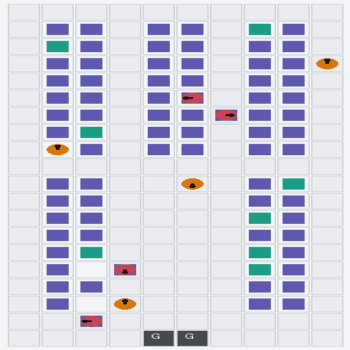}
        \caption{RWARE}
        \label{fig:rware_ss}
    \end{subfigure}
    \end{minipage}
\caption{\textbf{Evaluation benchmarks.} We evaluate PC3D on Spread, LBF, and RWARE, adapting each benchmark to episodic roster variation under fixed local observation interfaces.}
\label{fig:benchmarks}
\end{figure}

\paragraph{Benchmarks.} We evaluate PC3D on three fully cooperative MARL environments (Figure \ref{fig:benchmarks}). In each benchmark, the roster size varies across episodes, execution is decentralized, agents are homogeneous, and each agent acts based on its local observation history. We modify environments to use fixed-width local observations where possible, limiting exposure to trivial roster cues arising from changing observation dimensionality across roster sizes. The roster sizes we use in our training and evaluations are split into explicit training (seen during training), validation (unseen but used for selection during hyperparameter search, reserved for intermediate values), and held-out test counts (unseen and used only for reporting, reserved for larger counts to demonstrate extrapolation).

\textbf{Simple Spread} \cite{spread1} is a standard MPE particle-world coverage task, with a two-dimensional arena in which agents must spread out to cover the landmarks while avoiding collisions (Figure \ref{fig:spread_ss}). Our version was built from PettingZoo's \cite{pettingzoo} \texttt{simple\_spread\_v3} environment with discrete actions and $n$ landmarks for each $n$ agent roster. We modify the environment interface with shared team rewards (negative-sum of distances between each landmark and the closest respective agent minus collision penalties), disabled communication channels, and fixed-width local observations (retaining only the agent's own velocity and position, the three nearest landmarks and teammates). We use training roster sizes $\{1,2,4,6,8\}$, validation roster sizes $\{3,5,7\}$, and held-out test roster sizes $\{9,10\}$.

\textbf{Level-based foraging (LBF)}
\cite{lbf1, benchmarking_marl} is a grid-world mixed cooperative-competitive game (Figure \ref{fig:lbf_ss}) where agents and food items have levels and a food item can be collected only when adjacent agents execute the loading action with a sufficient combined level. We use the cooperative variant \texttt{Foraging-2s-10x10-\{n\}p-\{f\}f-coop-v3}, with sight range $2$. We report \textit{normalized team returns}, computed as the native team reward divided by the active roster size. We scale the number of food items (\texttt{f}) with the active roster size (\texttt{n}). We replace the native observation with a fixed-width local entity encoding that does not grow with team size. We use training roster sizes $\{2,4,6\}$, validation roster sizes $\{3,5\}$, and held-out test roster sizes $\{7,8\}$.

\textbf{Multi-Robot Warehouse (RWARE)} \cite{benchmarking_marl} is a robotic warehouse control benchmark in which robots move through aisles, pick up requested shelves, and deliver them to goal cells (\texttt{G}) (Figure \ref{fig:rware_ss}). We use the \texttt{rware-small-\{n\}ag-v2} layout ($20 \times 10$). The reward type is set to \texttt{global}, so every robot receives the same reward when the team successfully delivers the requested shelves. RWARE is sparse and congestion-sensitive: larger teams can increase throughput, but they also congest the passages and interfere with shelf retrieval. We use training roster sizes $\{2,4,6,8\}$, validation roster sizes $\{3,5,7\}$, and held-out test roster sizes $\{9,10\}$. 

\paragraph{Baselines.}
We compare PC3D against three MARL baselines chosen to evaluate its contribution under the same decentralized execution setting: agents act from local histories without execution-time communication, privileged coordinators, global observations, or problem-specific state decompositions. Our analyzes systematically evaluate the gains introduced by personalized context distillation and adaptive context use in isolation, rather than reporting an exhaustive benchmarking study. \textbf{IPPO} is the MARL adaptation of the Proximal Policy Optimization algorithm with an independent learning setting, where both training and execution are fully decentralized \cite{ippo, mappo}. \textbf{MAPPO} extends it with a centralized critic and serves as our backbone method \cite{mappo}. In our variable-roster setting, MAPPO uses a fixed-width critic input based on the maximum admitted roster size, with inactive slots masked. \textbf{PIC{\scriptsize-MAPPO}} replaces the fixed-width centralized critic with a permutation-invariant set critic \cite{liu_pic}, but it does not distill personalized teacher contexts or adaptively condition the actor on the recovered context. All method implementations employ recurrent and shared-parameter actor networks so that the same policy can be reused across roster sizes.

\paragraph{Training and evaluation.}
We adopt a staged curriculum that gradually increases roster diversity \cite{curriculum, transferable_akshat}. We specify these stages in Figure \ref{fig:curves}. We report five seeded repetitions for each method-task pair. For the results reported in Tables \ref{tab:all_results} and \ref{tab:all_ablations}, and Figure \ref{fig:boxes}, the \textbf{final checkpoints} are evaluated on train, validation, and test agent-count splits, with $100$ rollouts per count. Appendix \ref{sec:additional_results} provides additional supporting results, and Appendix \ref{sec:repro} details our implementations and experimental setting.

\subsection{Main results}
\label{sec:main_results}

Table \ref{tab:all_results} reports final-checkpoint performance across the three benchmarks. PC3D{\scriptsize-MAPPO} obtains the strongest mean return in all tasks and splits, including held-out roster sizes that are never seen during training. The gains are clearest in Spread, where the PC3D actor improves substantially over the second-best method (PIC{\scriptsize-MAPPO}). LBF shows the same pattern on a different reward scale, with PC3D improving validation and test returns while preserving better training performance. RWARE results appear noisier (reflected in larger standard deviations), but display the same pattern: PC3D improves its backbone (MAPPO) by a clear margin and performs the best on train, validation, and test counts.

\begin{table}[h]
\centering
\caption{\textbf{Evaluation performance across roster splits.} Returns (means $\pm$ standard deviations) across five seeded final checkpoints. For each seed, the mean is the average per-count evaluation returns within the corresponding train, validation, or test roster sizes. Higher is better for all tasks. LBF values are multiplied by $10^2$ for readability. \textbf{Bold} indicates the best method in each column.}
\label{tab:all_results}
\resizebox{\textwidth}{!}{%
\begin{tabular}{@{}lccc|ccc|ccc@{}}
\multirow{2}{*}{Method} & \multicolumn{3}{c}{Spread} & \multicolumn{3}{c}{LBF ($\times 1e2$)}   & \multicolumn{3}{c}{RWARE} \\ 
\cmidrule(lr){2-4}\cmidrule(lr){5-7}\cmidrule(l){8-10}
& Train  & Validation & Test & Train & Validation & Test & Train & Validation & Test \\ \midrule
IPPO &
$-57.06$ $\pm$ $1.6$ & $-65.01$ $\pm$ $2.0$ & $-103.43$ $\pm$ $2.6$ &
$6.84$ $\pm$ $1.4$ & $3.98$ $\pm$ $0.7$ & $8.24$ $\pm$ $0.6$ &
$2.58$ $\pm$ $1.5$ & $2.58$ $\pm$ $1.5$ & $6.17$ $\pm$ $3.0$ \\
MAPPO &
$-42.45$ $\pm$ $0.8$ & $-51.96$ $\pm$ $1.3$ & $-86.13$ $\pm$ $1.7$ &
$5.61$ $\pm$ $0.6$ & $3.44$ $\pm$ $0.5$ & $7.91$ $\pm$ $0.8$ &
$1.07$ $\pm$ $0.6$ & $0.99$ $\pm$ $0.6$ & $2.77$ $\pm$ $1.6$ \\
PIC{\scriptsize-MAPPO} &
$-42.00$ $\pm$ $0.5$ & $-50.34$ $\pm$ $0.9$ & $-84.42$ $\pm$ $1.3$ &
$6.84$ $\pm$ $1.4$ & $3.76$ $\pm$ $0.7$ & $8.40$ $\pm$ $1.1$ &
$2.30$ $\pm$ $1.4$ & $2.22$ $\pm$ $1.4$ & $5.67$ $\pm$ $2.8$ \\
PC3D{\scriptsize-MAPPO} &
\bm{$-39.90$ $\pm$ $0.7$} & \bm{$-48.09$ $\pm$ $1.0$} & \bm{$-79.18$ $\pm$ $1.5$} &
\bm{$7.91$ $\pm$ $0.7$} & \bm{$4.47$ $\pm$ $0.3$} & \bm{$8.98$ $\pm$ $0.1$} &
\bm{$3.58$ $\pm$ $1.5$} & \bm{$3.53$ $\pm$ $1.5$} & \bm{$7.73$ $\pm$ $2.7$} \\
\bottomrule
\end{tabular}
}
\end{table}

\begin{figure}[h]
    \centering
    \begin{subfigure}{0.32\linewidth}
        \includegraphics[width=\linewidth]{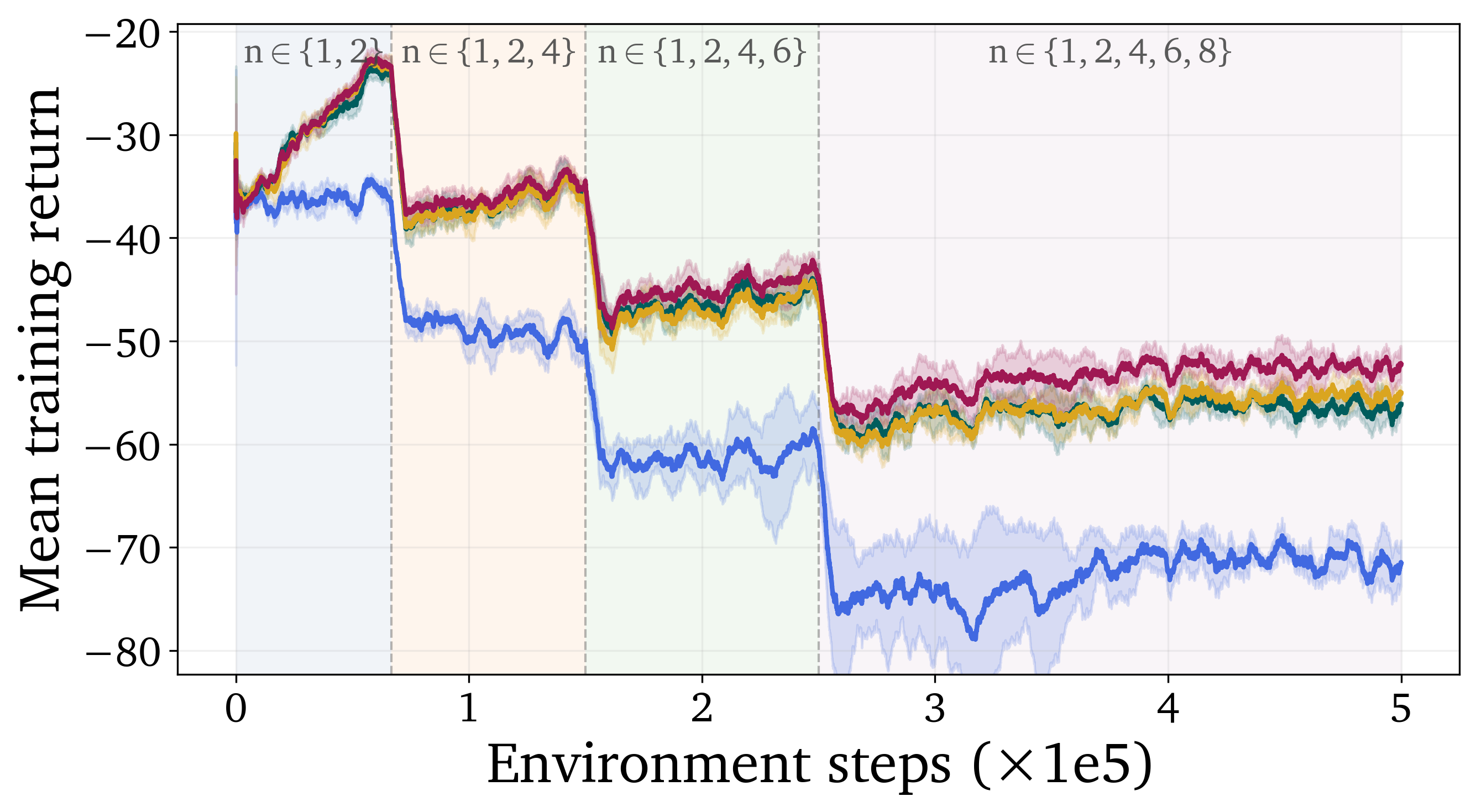}
        \caption{Spread}
        \label{fig:spread_curve}
    \end{subfigure}
    \hfill
    \begin{subfigure}{0.32\linewidth}
        \includegraphics[width=\linewidth]{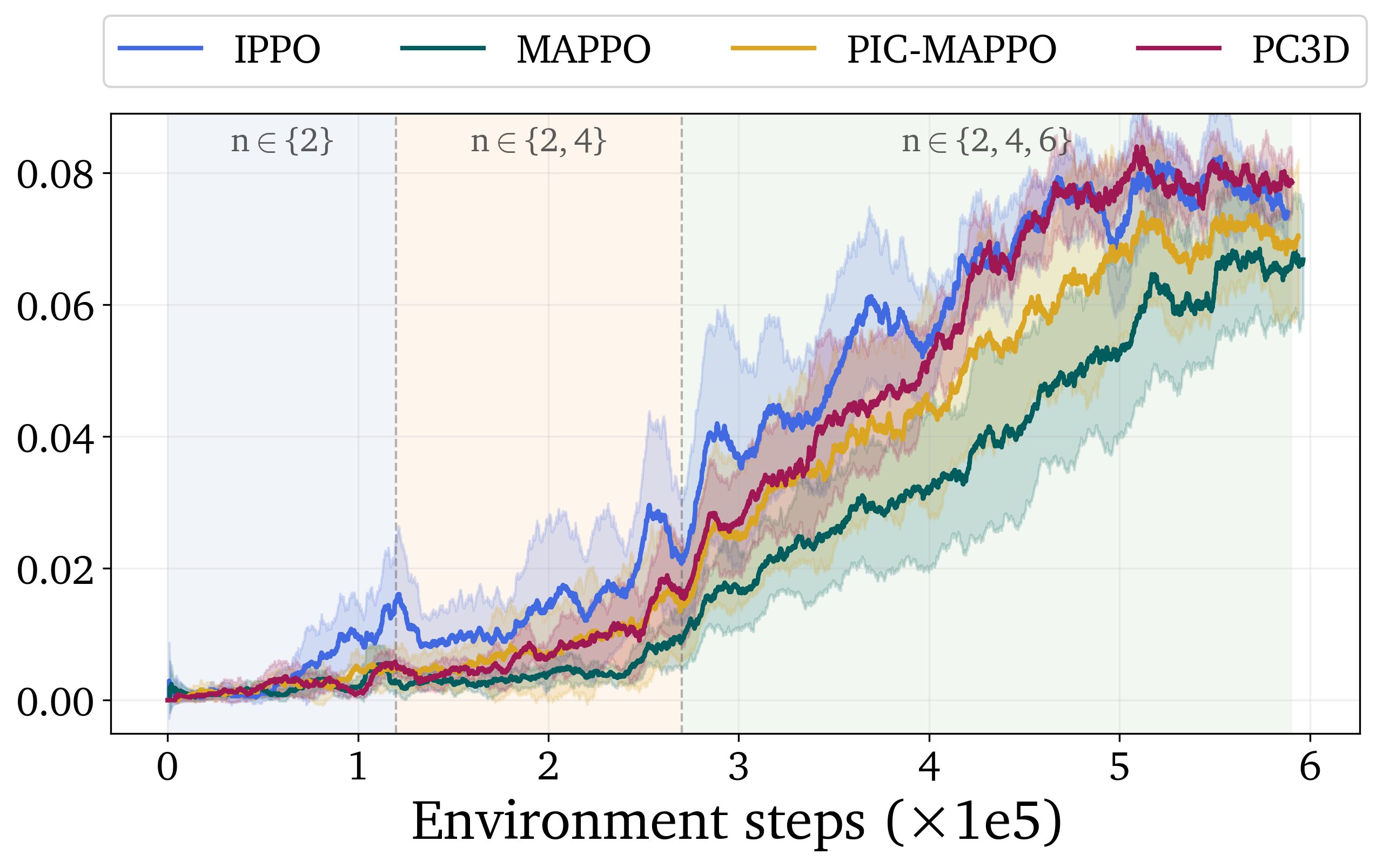}
        \caption{LBF}
        \label{fig:lbf_curve}
    \end{subfigure}
    \hfill
    \begin{subfigure}{0.32\linewidth}
        \includegraphics[width=\linewidth]{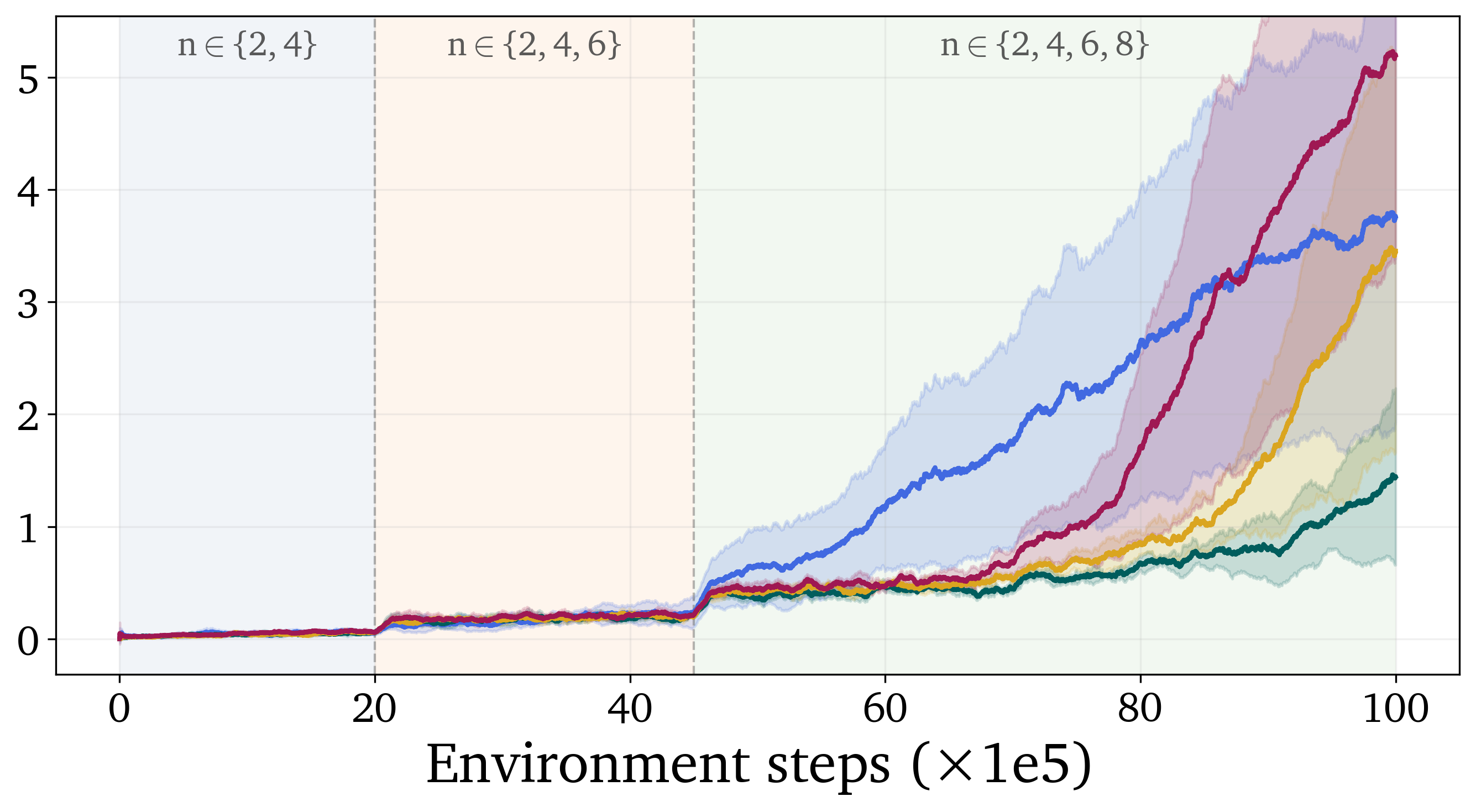}
        \caption{RWARE}
        \label{fig:rware_curve}
    \end{subfigure}
    \caption{\textbf{Training returns.} Curves show mean training returns ($\pm 95 \text{ CI}$) across seeds, with each colored patch corresponding to one curriculum stage and its active training roster set. Higher returns are better.}
    \label{fig:curves}
\end{figure}

The learning curves in Figure \ref{fig:curves} show optimization behavior under the active curriculum distribution. First, PC3D remains competitive throughout the curriculum, with more notable improvements over baselines in later stages as rosters grow and roster distribution becomes more diverse. This is consistent with the primary objective of PC3D: a method that extends the single-roster optimizers to generalize across diverse roster distributions.

\begin{figure}[h]
    \centering
    \begin{subfigure}{1\linewidth}
        \centering
        \includegraphics[width=\linewidth]{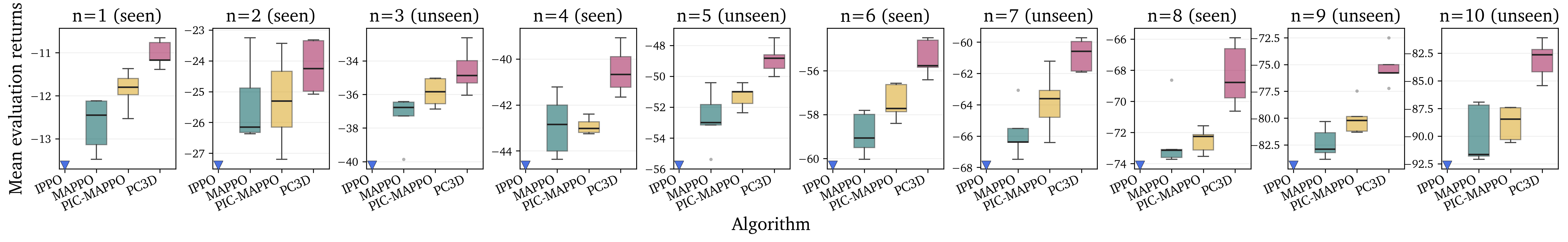}
        \caption{Spread}
        \vspace{6pt}
        \label{fig:spread_box}
    \end{subfigure}
    
    \begin{subfigure}{\linewidth}
        \centering
        \includegraphics[width=0.7\linewidth]{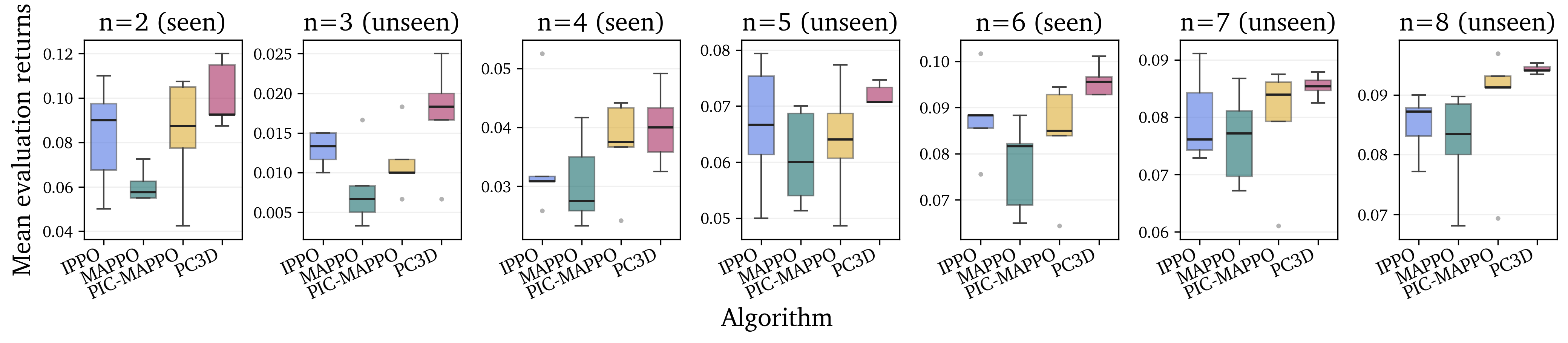}
        \caption{LBF}
        \vspace{6pt}
        \label{fig:lbf_box}
    \end{subfigure}
    
    \begin{subfigure}{\linewidth}
        \centering
        \includegraphics[width=0.9\linewidth]{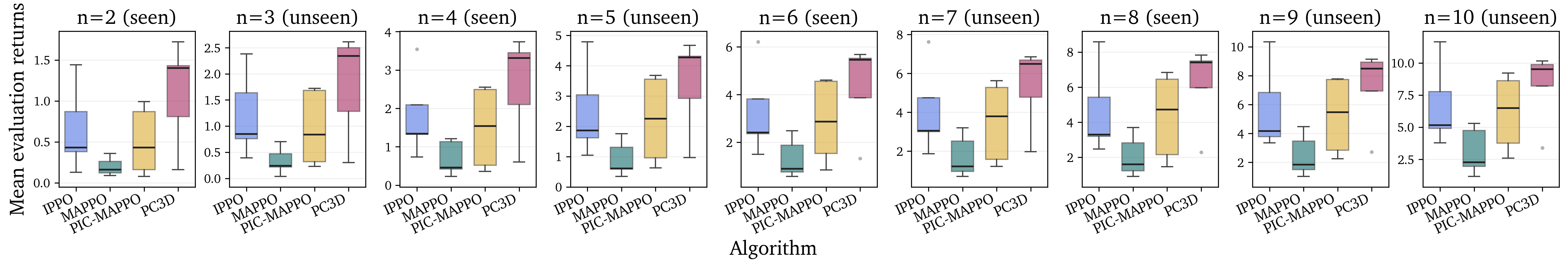}
        \caption{RWARE}
        \label{fig:rware_box}
    \end{subfigure}
    \caption{\textbf{Evaluation returns across roster sizes.} Final-checkpoint returns for each evaluated roster size. Each count is evaluated separately with $100$ rollouts. Downward markers indicate methods whose returns fall below the displayed range.}
    \label{fig:boxes}
\end{figure}

Figure \ref{fig:boxes} shows the evaluation returns for each method and benchmark across the used roster sizes. These plots better highlight count-specific performances that are compressed in the split means we report in Table \ref{tab:all_results}. PC3D generally shifts the return distribution upward across both seen and unseen counts, rather than improving only a single favorable roster size. In particular, evaluations on larger rosters show that PC3D widens the margin over baselines as coordination becomes less trivial.

\subsection{Ablations}
\label{sec:ablations}

We perform ablations to test whether the gains reported in Section \ref{sec:main_results} are correlated with our methodological objectives. Using the same PC3D runs (from Section \ref{sec:main_results}), we ablate the two mechanisms that form the basis of the intuition behind PC3D: context distillation and adaptive context conditioning. \textbf{Always Off Gate} sets $g_i^t=0$ (see Eq. \ref{eq:condition}), preventing context modulation; \textbf{Always On Gate} sets $g_i^t=1$, forcing non-adaptive modulation; and \textbf{A-MAPPO} sets $\lambda_{\mathrm{distill}}=0$ (see Eq. \ref{eq:final_loss}), retaining the attention critic and feature conditioning path without teacher-student alignment. The ablations use the same training and evaluation protocol as the corresponding PC3D runs.

\begin{table}[h]
\centering
\caption{\textbf{Ablation study.} We train three versions of each model with ablations. Entries report mean $\pm$ standard deviation across five seeded final checkpoints, using the same split-level aggregation as Table \ref{tab:all_results}. PC3D{\scriptsize-MAPPO} row is taken from Table \ref{tab:all_results}. \textbf{Bold} indicates the best variant in each column.}
\label{tab:all_ablations}
\resizebox{\textwidth}{!}{%
\begin{tabular}{@{}lccc|ccc|ccc@{}}
\multirow{2}{*}{Method} & \multicolumn{3}{c}{Spread} & \multicolumn{3}{c}{LBF ($\times 1e2$)}   & \multicolumn{3}{c}{RWARE} \\ 
\cmidrule(lr){2-4}\cmidrule(lr){5-7}\cmidrule(l){8-10}
& Train  & Validation & Test & Train & Validation & Test & Train & Validation & Test \\ \midrule
PC3D{\scriptsize-MAPPO} &
$-39.90$ $\pm$ $0.7$ & \bm{$-48.09$ $\pm$ $1.0$} & \bm{$-79.18$ $\pm$ $1.5$} &
\bm{$7.91$ $\pm$ $0.7$} & $4.47$ $\pm$ $0.3$ & \bm{$8.98$ $\pm$ $0.1$} & 
\bm{$3.58$ $\pm$ $1.5$} & \bm{$3.53$ $\pm$ $1.5$} & \bm{$7.73$ $\pm$ $2.7$} \\
Always Off Gate & 
$-41.71$ $\pm$ $0.8$ & $-50.23$ $\pm$ $0.7$ & $-82.95$ $\pm$ $2.3$ & 
$7.09$ $\pm$ $0.5$ & $3.54$ $\pm$ $0.6$ & $8.39$ $\pm$ $0.5$ & 
$2.14$ $\pm$ $0.9$ & $2.00$ $\pm$ $0.9$ & $5.43$ $\pm$ $2.1$ \\
Always On Gate & 
$-41.40$ $\pm$ $0.7$ & $-49.67$ $\pm$ $1.1$ & $-82.20$ $\pm$ $2.1$ & 
$7.67$ $\pm$ $1.0$ & $4.28$ $\pm$ $0.5$ & $8.93$ $\pm$ $0.3$ & 
$3.19$ $\pm$ $2.1$ & $3.06$ $\pm$ $2.0$ & $7.22$ $\pm$ $3.9$ \\
A-MAPPO & 
\bm{$-39.78$ $\pm$ $0.9$} & $-48.24$ $\pm$ $1.1$ & $-80.65$ $\pm$ $2.2$ & 
$7.54$ $\pm$ $0.7$ & \bm{$4.61$ $\pm$ $0.4$} & $8.97$ $\pm$ $0.4$ & 
$2.69$ $\pm$ $1.7$ & $2.58$ $\pm$ $1.7$ & $6.14$ $\pm$ $2.9$ \\  
\bottomrule
\end{tabular}
}
\end{table}

The results presented in Table \ref{tab:all_ablations} support three conclusions. First, turning the gate off consistently hurts performance, showing that the learned context pathway is not a passive auxiliary head (also supported in Appendix \ref{sec:recovery_use}). Second, forcing the gate on is competitive in LBF but notably weaker in Spread and RWARE, suggesting that adaptive reliance is most useful when roster diversity increases and task demands vary across roster sizes. Third, removing distillation can occasionally remain competitive (most notably Spread seen and LBF unseen splits), but it weakens generalization in Spread and substantially hurts RWARE across splits. Overall, these results suggest that the centralized teacher does not merely improve the critic; it provides a personalized signal that helps the recurrent actor recover and use coordination context under roster shift.

%% file: sections/05_conclusions.tex
\section{Conclusions}
\label{sec:conclusions}

This study focused on open-team cooperation under episodic roster variation and partial observability, where fully decentralized agents cooperate across varying and unseen team sizes. We formalized this setting as a family of roster-indexed Dec-POMDPs induced by a shared template, and argued that standard CTDE methods lack an explicit mechanism for turning centralized coordination information into a reusable decentralized representation.

We introduced \textbf{PC3D} on top of a MAPPO backbone as a method that trains a set-structured centralized teacher to personalize its context and distill it into decentralized policies. The resulting actor recovers a student coordination context from local history and adaptively uses it through gated feature modulation. In contrast to approaches that address OTC by introducing structural assumptions, PC3D preserves the fully decentralized execution contract while providing the policy with a direct training signal to recover useful coordination context from local interactions, supporting zero-shot adaptation across varying roster sizes.

Across Spread, LBF, and RWARE, PC3D improves over IPPO, MAPPO, and PIC{\scriptsize-MAPPO} on both seen and unseen roster sizes. Furthermore, the ablations support that non-adaptive modulation or the removal of distillation weakens performance, especially under larger roster shifts. Generally, our results indicate that open-team cooperation should be treated not only as a robustness problem but also as a representation-transfer problem between centralized training and decentralized execution.

Extending PC3D to heterogeneous teams and in-episode roster changes is a natural next step. Moreover, testing it with value-factorization or off-policy critic CTDE methods (see Appendix \ref{sec:discussion}) would clarify the transferability of the coordination-distillation principle. PC3D is least compelling when execution permits communication or centralized observations, or when the task does not contain a reusable cooperative structure across rosters. It is intended for settings where centralized roster-dependent representations can be personalized and meaningfully guide decentralized execution. PC3D aims to support more robust decentralized coordination in robotics, logistics, and distributed control. However, deployment in safety-critical settings requires additional validation, as failures in unseen team configurations could lead to unsafe collective behavior.

%% file: sections/a_additional_results.tex
\section{Additional results}
\label{sec:additional_results}

\subsection{Training returns}

\begin{figure}[ht]
    \centering
    \begin{subfigure}{0.93\linewidth}
        \includegraphics[width=\linewidth]{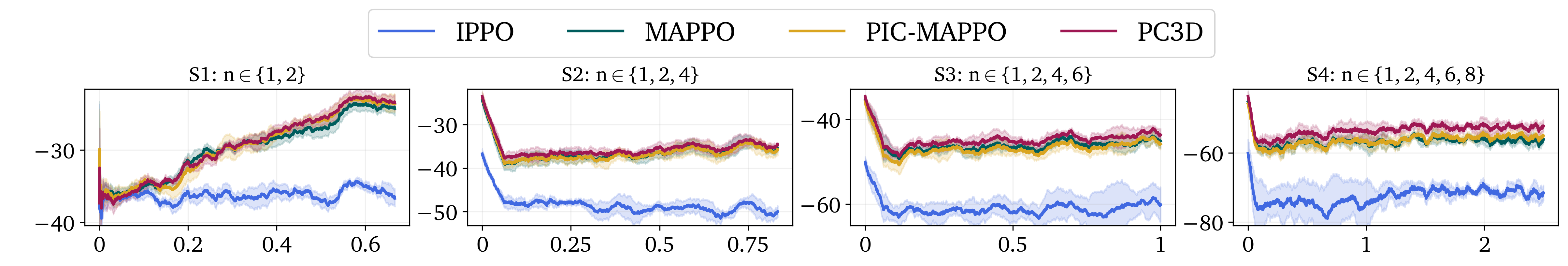}
        \caption{Spread}
        \vspace{6pt}
        \label{fig:spread_per_stage_curve}
    \end{subfigure}
    
    \begin{subfigure}{\linewidth}
        \includegraphics[width=\linewidth]{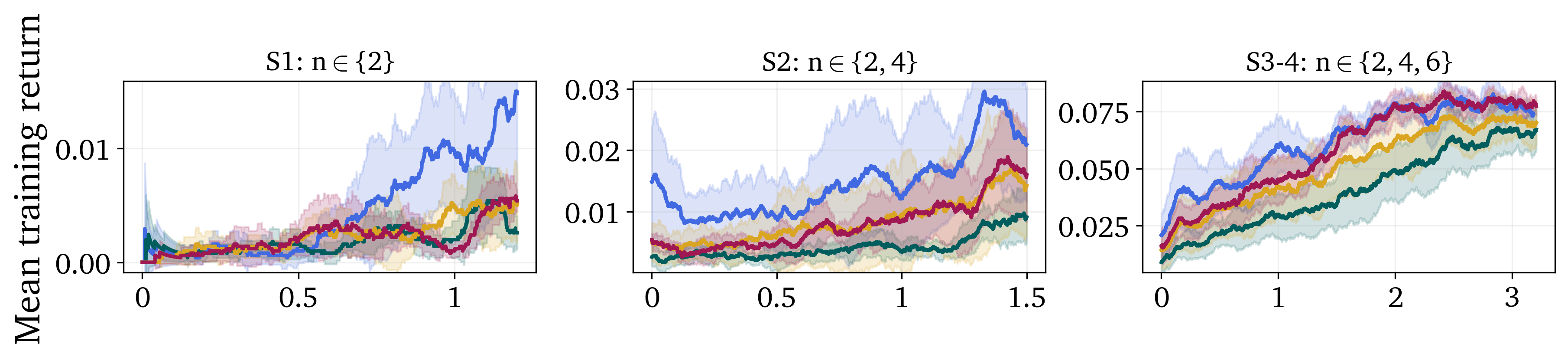}
        \caption{LBF}
        \vspace{6pt}
        \label{fig:lbf_per_stage_curve}
    \end{subfigure}
    
    \begin{subfigure}{\linewidth}
    \includegraphics[width=\linewidth]{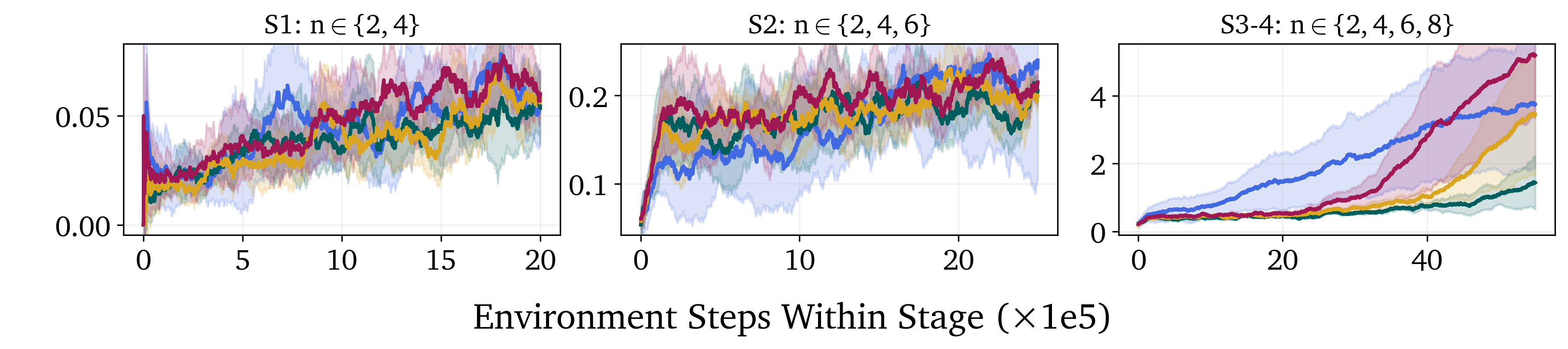}
        \caption{RWARE}
        \label{fig:rware_per_stage_curve}
    \end{subfigure}
    \caption{\textbf{Training curves across curriculum stages.} Curves show mean training returns across seeds, with each panel corresponding to one curriculum stage and its active training roster set. Stage-separated plots highlight the perturbations with increasing roster diversity. Higher returns are better.}
    \label{fig:per_stage_curves}
\end{figure}

Figure \ref{fig:per_stage_curves} displays the mean training returns across repetitions, using subplots for each curriculum stage. This visualization displays the same data as in Figure \ref{fig:curves}, but it is intended to highlight stage-wise behavior without compressing all stages onto a shared reward scale.

\subsection{Context recovery and use}
\label{sec:recovery_use}

\begin{figure}[ht]
    \centering
    \begin{subfigure}{1\linewidth}
        \includegraphics[width=\linewidth]{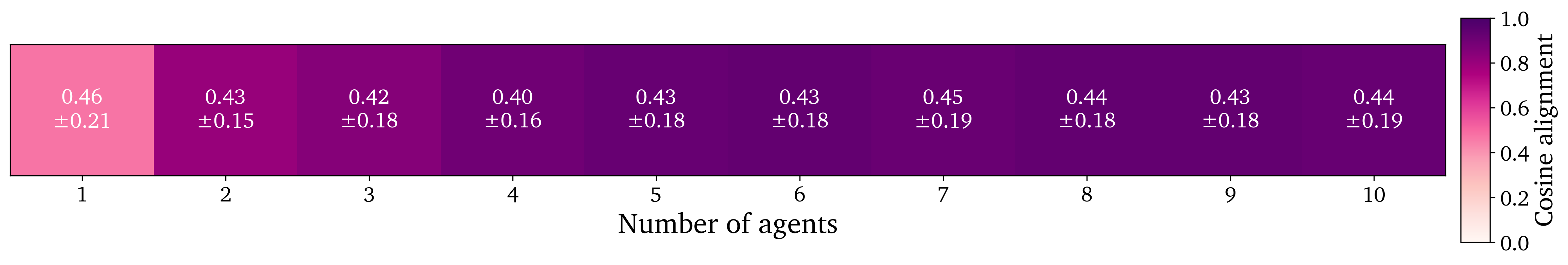}
        \caption{Spread}
        \vspace{6pt}
        \label{fig:spread_subalign}
    \end{subfigure}
    
    \begin{subfigure}{0.75\linewidth}
        \includegraphics[width=\linewidth]{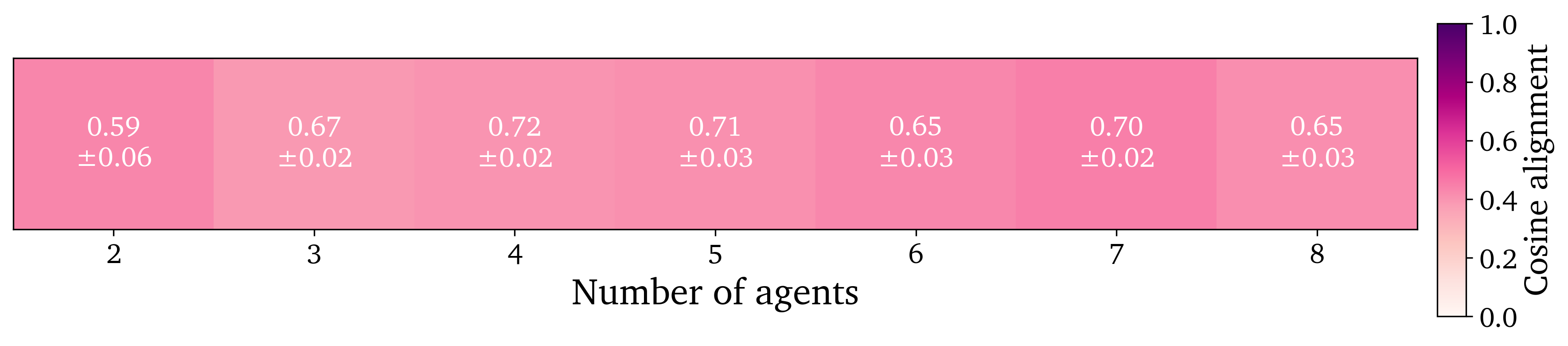}
        \caption{LBF}
        \vspace{6pt}
        \label{fig:lbf_subalign}
    \end{subfigure}
    
    \begin{subfigure}{0.9\linewidth}
        \includegraphics[width=\linewidth]{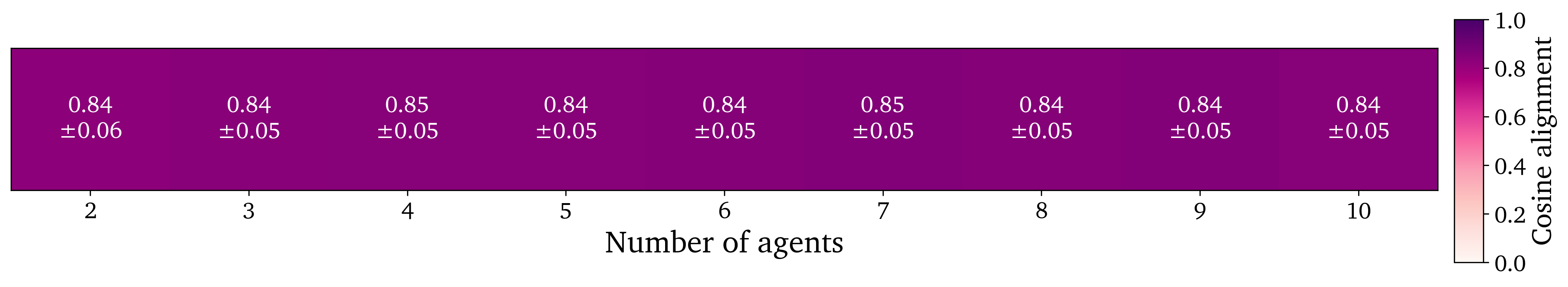}
        \caption{RWARE}
        \label{fig:rware_subalign}
    \end{subfigure}
    \caption{\textbf{Context recovery and use.} We run reported final checkpoints for 8 rollouts for each task and roster size pair. Cell color shows teacher-student cosine alignment for the personalized coordination context; cell text shows mean $\pm$ standard deviation of the context-reliance gate across repetitions. The plots test whether PC3D's centralized context is both locally recoverable and adaptively used by the decentralized actor.}
    \label{fig:full_alignment}
\end{figure}

Although the results reported in Section \ref{sec:main_results} show that PC3D improves policy performance, the distillation objective is compelling only if the decentralized actor actually recovers a useful teacher signal from local history. The findings reported in Table \ref{tab:all_ablations} are useful, but we probe this mechanism more directly in Figure \ref{fig:full_alignment} by reporting the context recovery and adaptive gating of the PC3D models (reported in Table \ref{tab:all_results}). Each cell corresponds to one evaluated roster size. High alignment provides evidence that the decentralized recurrent actor can reconstruct the centralized personalized context; nonzero gate values indicate that the modulation path is active rather than trivially suppressed; and deviations from the mean gate indicate the adaptiveness of policy conditioning.

\subsection{Policy conditioning using hypernetworks}
\label{sec:hyper}

Although FiLM modulation-based policy conditioning provides a clear and interpretable structure for local context use, one could achieve similar functionality using \textit{hypernetworks} \cite{hypernetworks, hypernets_review}. Hypernetworks have been previously adopted in MARL \cite{botteghi_hype}, especially for behavioral diversity in shared-parameter settings \cite{tessera_hyper}. In this section, we report additional results obtained with another version of our method that we have tested, with hypernetwork-based policy adaptation (\textit{Hyper-PC3D{\scriptsize-MAPPO}}) in two benchmarks. Specifically, this version replaces the transformations in Equation \ref{eq:condition} with:

\begin{equation} \label{eq:condition_v7}
(\Delta W_i^t, \Delta b_i^t) = H_\eta(\hat c_i^t),
\qquad
\ell_i^t
=
(h_i^t)^\top
\left(W_0 + g_i^t \Delta W_i^t\right)
+
\left(b_0 + g_i^t \Delta b_i^t\right),
\end{equation}

where $H_\eta$ is a context-conditioned hypernetwork that predicts a residual adaptation of the policy head, and $W_0$ and $b_0$ are the base trainable policy parameters. Thus, instead of FiLM-modulating the recurrent feature $h_i^t$, this variant keeps $h_i^t$ unchanged and uses the recovered context to adapt the mapping from recurrent policy features to action logits, while the context-reliance gate $g_i^t$ controls the strength of this policy-head adaptation.

\begin{table}[ht]
\centering
\caption{\textbf{Comparison of PC3D{\scriptsize-MAPPO} and Hyper-PC3D{\scriptsize-MAPPO} on two benchmarks.} Returns (means $\pm$ standard deviations) across five seeded final checkpoints. For each seed, the mean is the average per-count evaluation returns within the corresponding train, validation, or test roster sizes. Higher is better for all tasks. LBF values are multiplied by $10^2$ for readability. \textbf{Bold} indicates the better result.}
\label{tab:v6_v7_comparison}
\resizebox{\textwidth}{!}{%
\begin{tabular}{@{}lccc|ccc@{}}
\toprule
\multirow{2}{*}{Method} & \multicolumn{3}{c}{Spread} & \multicolumn{3}{c}{LBF ($\times 10^2$)} \\ 
\cmidrule(lr){2-4}\cmidrule(lr){5-7}
 & Train & Validation & Test & Train & Validation & Test \\ \midrule
PC3D{\scriptsize-MAPPO} & $-39.90$ $\pm$ $0.72$ & \bm{$-48.09$ $\pm$ $1.03$} & \bm{$-79.18$ $\pm$ $1.52$} & \bm{$7.91$ $\pm$ $0.70$} & \bm{$4.47$ $\pm$ $0.30$} & \bm{$8.98$ $\pm$ $0.08$} \\
Hyper-PC3D{\scriptsize-MAPPO} & \bm{$-39.62$ $\pm$ $1.00$} & $-48.41$ $\pm$ $1.40$ & $-80.58$ $\pm$ $1.95$ & $6.73$ $\pm$ $1.02$ & $3.97$ $\pm$ $0.61$ & $8.54$ $\pm$ $0.63$ \\
\bottomrule
\end{tabular}
}
\end{table}

The results show that, in the tested benchmarks, conditioning the recurrent policy features is more effective than adapting the policy head itself. While the hypernetwork-based approach is competitive in Spread, it shows a clear degradation in LBF. A plausible explanation is that FiLM modulation lets the recovered coordination context reshape the actor's internal representation while preserving a stable mapping. In contrast, the hypernetwork variant uses the context to generate residual changes to the policy-head parameters, which is more expressive but less constrained and may encourage roster-specific adaptations that transfer less reliably to held-out rosters. Therefore, we concluded that PC3D benefits more from the recovered context as a feature-level adaptation signal than as a direct policy-head adaptation signal, motivating us to base our study on the FiLM-based version. However, we note that different applications with larger roster spaces may require more adaptive representations to handle drastically different cooperation regimes caused by large fluctuations in roster sizes, which, in principle, could potentially be better addressed by the behavioral diversity that can be obtained with hypernetworks.

%% file: sections/b_novelty.tex
\section{Execution Assumptions in Related Work}
\label{sec:novelty}

Table \ref{tab:novelty} summarizes how representative variable-team CMARL methods (mentioned in Section \ref{sec:intro}) differ in their execution-time assumptions.

\begin{table}[h]
\centering
\caption{\textbf{Execution time assumptions in related variable-team MARL settings.}
We compare representative methods addressing CMARL with dynamic team sizes in terms of what they require at execution time. The comparison is intended to clarify the setting addressed in this work: PC3D preserves the standard CTDE contract with fully decentralized execution, with each agent acts from its own local history, without communication, privileged execution-time information, or additional designer-specified structure.}
\label{tab:novelty}
\resizebox{\textwidth}{!}{%
\small
\newcommand{\cmark}{\checkmark}
\newcommand{\xmark}{--}
\begin{tabular}{@{}lccccc@{}}
\toprule
Method
& Central module
& Communication
& Privileged info.
& Full DE
& Additional assumptions \\
\midrule
Agent--entity graphs \cite{transferable_akshat}
& \xmark
& \cmark
& \xmark
& \xmark
& Entity graph/message structure \\
COPA \cite{copa}
& \cmark
& \cmark
& \cmark
& \xmark
& Omniscient coach \\
SOG \cite{sog}
& \xmark
& \cmark
& \xmark
& \xmark
& \xmark \\
MIPI \cite{mipi}
& \xmark
& \xmark
& \xmark
& \cmark
& Designer-defined $s^+/s^-$ split \\
PC3D (ours)
& \xmark
& \xmark
& \xmark
& \cmark
& \xmark \\
\bottomrule
\end{tabular}}
\end{table}

%% file: sections/c_discussion.tex
\section{Discussion: PC3D as a CTDE extension}
\label{sec:discussion}

PC3D is designed as an extension for CTDE methods that train with centralized information but execute with local policies, rather than as a standalone optimizer itself. At a high level, it requires (i) a centralized module that can learn a value-relevant team representation, (ii) local features from which each agent can predict a student context, and (iii) an execution policy that can be conditioned on this recovered context without using centralized information.

This makes PC3D the most naturally adaptable for MAPPO or HAPPO-style \cite{mappo, happo} stochastic actor–critic methods. In such methods, the centralized critic can be extended with the PC3D set teacher, while a student-context head and a gated FiLM conditioning path are added to the decentralized actor. The policy objective would remain the original actor-critic objective; PC3D only adds a teacher-student context distillation loss and a context-conditioned actor representation.

Although off-policy centralized-critic methods such as MADDPG \cite{maddpg}, MATD3 \cite{matd3}, and MASAC \cite{benchmarl} are compatible in principle, they require additional method-specific adaptation because their critics estimate action values rather than state values. In a PC3D-style extension, the distilled teacher context should be constructed from agent observations before joint actions are introduced, while the realized joint action could be used only in the downstream action-value head. However, if the teacher context depends directly on simultaneous teammate actions, the decentralized student may not be able to reconstruct it from local history alone, thereby weakening the recoverability principle on which PC3D relies.

PC3D can also be theoretically compatible with QMIX-style value-factorization methods \cite{qmix}. In this setting, the mixer and its state-conditioned hypernetwork can provide the centralized training signal needed for extracting team-level coordination context. The joint TD loss can shape the teacher module to extract value-relevant coordination context, while in parallel, each recurrent utility network learns a student context from local history and uses it to condition the features before the per-agent value head. The only structural constraint is that the mixer remains monotonic with respect to the per-agent utilities. PC3D may condition each $Q_i$ through the recovered student context and may feed the teacher summary into the mixer hypernetworks, but the mixer weights must remain constrained so that $\partial Q_{\mathrm{tot}}/\partial Q_i \ge 0$. Under this constraint, decentralized greedy action selection remains valid, and PC3D becomes a context-distillation extension of the factorized value learner rather than an actor-critic-specific mechanism.

Finally, in our descriptions, we assumed that a separate global state is not available and constructed the centralized critic input from agent observations. In case a global state is available, it can be appended to the value head input while keeping the personalized teacher-context pathway observation-based, preserving the intended link between teacher personalization and local recoverability. This could improve value prediction, but may entail a trade-off of weakening the teacher-token learning signal if the critic relies on state features directly.

%% file: sections/d_repro.tex
\section{Reproducibility}
\label{sec:repro}

This section includes our implementation and parameterization details to support the reproducibility of this research.

\subsection{Hyperparameters}
\label{sec:hyperparams}

Table \ref{tab:hyperparams} reports the hyperparameters used for training the models evaluated in Section \ref{sec:main_results}. All models are trained with hyperparameters selected as performant (on training and validation splits) after a hyperparameter search for each algorithm-benchmark pair. To conduct a fair comparison, \textbf{the hyperparameters used for MAPPO are reused by PIC{\scriptsize-MAPPO} and PC3D} (except for the critic head shape, as the inputs and their shapes differ substantially), while IPPO is tuned independently.

\begin{table}[ht]
\centering
\scriptsize
\setlength{\tabcolsep}{3pt}
\caption{\textbf{Final evaluation hyperparameters.} Configurations used for the models reported in Section \ref{sec:main_results}. Rows marked with $^\dagger$ denote parameters chained from the same-task MAPPO finalist into PIC{\scriptsize-MAPPO} and PC3D. For boolean entries, T/F indicates whether the option is enabled.}
\label{tab:hyperparams}
\resizebox{\textwidth}{!}{%
\begin{tabular}{@{}lcccccccccccc@{}}
\toprule
\multirow{2}{*}{Parameter}
& \multicolumn{4}{c}{Spread}
& \multicolumn{4}{c}{LBF}
& \multicolumn{4}{c}{RWARE} \\
\cmidrule(lr){2-5}\cmidrule(lr){6-9}\cmidrule(l){10-13}
& IPPO & MAPPO & PIC & PC3D
& IPPO & MAPPO & PIC & PC3D
& IPPO & MAPPO & PIC & PC3D \\
\midrule
Optimizer 
& Adam & Adam & Adam & Adam 
& Adam & Adam & Adam & Adam
& Adam & Adam & Adam & Adam \\
Learning rate$^\dagger$
& 1.46e-4 & 1.84e-3 & 1.84e-3 & 1.84e-3
& 1.22e-3 & 4.63e-4 & 4.63e-4 & 4.63e-4
& 2.06e-4 & 1.24e-4 & 1.24e-4 & 1.24e-4 \\
Batch size$^\dagger$
& 128 & 128 & 128 & 128
& 128 & 256 & 256 & 256
& 64 & 64 & 64 & 64 \\
Buffer size
& 8192 & 200000 & 200000 & 200000
& 8192 & 200000 & 200000 & 200000
& 8192 & 200000 & 200000 & 200000 \\
Update every eps.$^\dagger$
& 2 & 8 & 8 & 8
& 16 & 2 & 2 & 2
& 8 & 1 & 1 & 1 \\
PPO epochs$^\dagger$
& 6 & 8 & 8 & 8
& 6 & 8 & 8 & 8
& 8 & 6 & 6 & 6 \\
Actor widths$^\dagger$
& [96,128,128,96] & [128,256,128] & [128,256,128] & [128,256,128]
& [128,256,128] & [64,64] & [64,64] & [64,64]
& [96,128,128,96] & [128,256,128] & [128,256,128] & [128,256,128] \\
RNN dim$^\dagger$
& 128 & 128 & 128 & 128
& 64 & 128 & 128 & 128
& 192 & 32 & 32 & 32 \\
Clip $\epsilon^\dagger$
& .25 & .15 & .15 & .15
& .25 & .25 & .25 & .25
& .10 & .25 & .25 & .25 \\
Discount $\gamma^\dagger$
& .99 & .985 & .985 & .985
& .99 & .985 & .985 & .985
& .97 & .99 & .99 & .99 \\
GAE $\lambda^\dagger$
& .99 & .99 & .99 & .99
& .95 & .93 & .93 & .93
& .97 & .97 & .97 & .97 \\
Entropy coef.$^\dagger$
& 6.61e-4 & 1.28e-3 & 1.28e-3 & 1.28e-3
& 8.24e-3 & 1.11e-2 & 1.11e-2 & 1.11e-2
& 2.49e-3 & 2.34e-4 & 2.34e-4 & 2.34e-4 \\
Value coef.$^\dagger$
& .5 & .25 & .25 & .25
& .25 & 2.0 & 2.0 & 2.0
& .25 & .5 & .5 & .5 \\
Max grad norm$^\dagger$
& .5 & 2.0 & 2.0 & 2.0
& 5.0 & 5.0 & 5.0 & 5.0
& 1.0 & 10.0 & 10.0 & 10.0 \\
Critic widths
& -- & [128,96] & [128,128] & [192,160]
& -- & [160,160] & [160,128] & [160,128]
& -- & [128,96] & [96,96] & [96,96] \\
Set embed dim
& -- & -- & 48 & 48
& -- & -- & 96 & 48
& -- & -- & 160 & 96 \\
Set encoder $\phi$ widths
& -- & -- & [160,96] & [96,96]
& -- & -- & [96,64] & [160,96]
& -- & -- & [48,48] & [96,96] \\
Team-size feature
& -- & -- & F & T
& -- & -- & T & T
& -- & -- & T & F \\
Teacher tokens count $K$
& -- & -- & -- & 4
& -- & -- & -- & 4
& -- & -- & -- & 5 \\
Distill weight $\lambda_{\text{distill}}$
& -- & -- & -- & .257
& -- & -- & -- & .0193
& -- & -- & -- & .0154 \\
Teacher EMA $\tau$
& -- & -- & -- & .02
& -- & -- & -- & .0025
& -- & -- & -- & .0025 \\
Reliance clip $[\rho_{\text{min}}, \rho_{\text{max}}]$
& -- & -- & -- & [-3,2]
& -- & -- & -- & [-2,1.5]
& -- & -- & -- & [-3,2] \\
\bottomrule
\end{tabular}
}
\end{table}

\subsection{Implementation details}
\label{sec:imp_details}

\paragraph{PIC{\scriptsize-MAPPO}.}
PIC{\scriptsize-MAPPO} is the permutation-invariant critic baseline used in Section \ref{sec:results} to disentangle the effect of PC3D from the effect of replacing MAPPO's fixed-width centralized critic. It keeps the same recurrent shared actor as MAPPO, but replaces the critic input concatenation over padded agent slots with a permutation-invariant team encoder. Our implementation was inspired by the idea introduced in a prior study \cite{liu_pic}, originally proposed to handle agent-order changes in the critic inputs on a MADDPG backbone, and is, by design, applicable to our problem. For active roster $r^t$, each local observation is embedded as $e_i^t=\phi_\psi(o_i^t)$ and the critic forms
\begin{equation}
\bar e^t=\frac{1}{|r^t|}\sum_{i\in r^t} e_i^t,
\qquad
V^t=\rho_\omega(\bar e^t).
\end{equation}
Thus, PIC{\scriptsize-MAPPO} gives MAPPO a permutation-invariant, variable-size-compatible centralized value function, but does not use coordination tokens, personalized teacher contexts, context distillation, or actor conditioning.

\paragraph{Baseline implementations.}
All reported methods use recurrent agent networks and parameter sharing across agents. IPPO uses a shared recurrent actor architecture with a local value head and no centralized critic. MAPPO uses a shared recurrent actor and a centralized critic over padded joint observations with active-agent masks. PIC{\scriptsize-MAPPO} replaces this fixed-width critic with the set critic described above. PC3D uses the same shared recurrent actor backbone but augments training with a permutation-invariant centralized teacher, personalized context distillation, and gated FiLM conditioning of the actor features. For PIC{\scriptsize-MAPPO} and PC3D{\scriptsize-MAPPO}, the centralized value head can optionally receive the active roster size as an additional scalar feature (the settings used in the reported runs are listed in Table \ref{tab:hyperparams}, by parameter \texttt{Team-size feature}). In all cases, execution is decentralized: each agent acts on its own observation history, without communication, global observations, or privileged state decompositions, to make our evaluation results more interpretable and less influenced by non-methodological factors.

\paragraph{Roster splits and curricula.}
Each task defines train, validation, and test roster sizes. During the hyperparameter search, we used only train and validation counts; test counts are reserved only for final reporting. During training, each episode samples one roster size from the current curriculum stage according to the probabilities listed in Table \ref{tab:curricula}.

\begin{table}[t]
\centering
\caption{\textbf{Training curricula.} Each episode samples one roster size from the active stage. Probabilities are listed in the same order as the roster counts.}
\label{tab:curricula}
\resizebox{\textwidth}{!}{%
\begin{tabular}{@{}lcccc@{}}
\toprule
Task & Stage & Episode fraction & Roster counts & Sampling probabilities \\
\midrule
\multirow{4}{*}{Spread}
& 1 & $13.3\%$ & $\{1,2\}$ & $(0.40,0.60)$ \\
& 2 & $16.7\%$ & $\{1,2,4\}$ & $(0.18,0.27,0.55)$ \\
& 3 & $20.0\%$ & $\{1,2,4,6\}$ & $(0.10,0.15,0.30,0.45)$ \\
& 4 & $50.0\%$ & $\{1,2,4,6,8\}$ & $(0.06,0.09,0.18,0.27,0.40)$ \\
\midrule
\multirow{4}{*}{LBF}
& 1 & $20.0\%$ & $\{2\}$ & $(1.00)$ \\
& 2 & $25.0\%$ & $\{2,4\}$ & $(0.35,0.65)$ \\
& 3 & $25.0\%$ & $\{2,4,6\}$ & $(0.15,0.25,0.60)$ \\
& 4 & $30.0\%$ & $\{2,4,6\}$ & $(0.10,0.20,0.70)$ \\
\midrule
\multirow{4}{*}{RWARE}
& 1 & $20.0\%$ & $\{2,4\}$ & $(0.65,0.35)$ \\
& 2 & $25.0\%$ & $\{2,4,6\}$ & $(0.30,0.20,0.50)$ \\
& 3 & $25.0\%$ & $\{2,4,6,8\}$ & $(0.10,0.15,0.25,0.50)$ \\
& 4 & $30.0\%$ & $\{2,4,6,8\}$ & $(0.05,0.10,0.20,0.65)$ \\
\bottomrule
\end{tabular}
}
\end{table}

\paragraph{Hyperparameter search and parameter chaining.}
We tune the MAPPO backbone first and reuse its actor/PPO hyperparameters for PIC{\scriptsize-MAPPO} and PC3D. This parameter chaining keeps comparisons focused on the architectural changes rather than allowing each method to compensate through unrelated PPO settings. The chained parameters include the learning rate, batch size, update frequency, PPO epochs, actor width/depth, recurrent hidden size, clipping coefficient, discounting, GAE parameter, entropy coefficient, value coefficient, and gradient clipping (see Table \ref{tab:hyperparams}). Method-specific parameters, such as the set-critic embedding size, critic hidden sizes, number of PC3D tokens, distillation weight, teacher moving-average rate, and context-reliance bounds, are tuned separately.

\paragraph{Final evaluations.}
After selecting configurations, we rerun each configuration with a larger training budget using five random seeds and report the final checkpoint, not the best validation checkpoint. The final evaluation uses $100$ rollouts per roster across train, validation, and test sets. For each seed and split, we first average returns within each roster count and then average across counts in that split; tables report the mean and standard deviation of these split-level values across seeds. Spread is trained for $20{,}000$ episodes, LBF for $12{,}000$ episodes, and RWARE for $20{,}000$ episodes. For Spread, the number of landmarks is always equal to the number of agents. For LBF, the number of food items is tied to roster size, with $\{2,3\}$ agents using $2$ food items, $\{4,5,6\}$ using $3$, and $\{7,8\}$ using $4$.

\subsection{Resources}
\label{sec:resources}

Table \ref{tab:compute_overhead} reports the average wall-clock time required to complete one learner update for each method and benchmark. Table \ref{tab:hardware_specs} summarizes the hardware and execution setting used for these measurements.

\begin{table}[ht]
\centering
\caption{\textbf{Wall clock time for policy updates.}
Values report seconds per learner update, computed as total training-run wall-clock time divided by the number of policy updates completed during a 512-episode CPU run.
}
\label{tab:compute_overhead}
\begin{tabular}{@{}lccc@{}}
\toprule
Method & Spread & LBF & RWARE \\
\midrule
IPPO & $3.88$ & $6.08$ & $23.73$ \\
MAPPO & $19.00$ & $43.00$ & $76.50$ \\
PIC{\scriptsize-MAPPO} & $16.75$ & $38.00$ & $73.88$ \\
PC3D{\scriptsize-MAPPO} & $20.25$ & $49.00$ & $95.00$ \\
\bottomrule
\end{tabular}
\end{table}

\begin{table}[ht]
\centering
\caption{\textbf{Hardware and execution settings used for values in Table \ref{tab:compute_overhead}.}}
\label{tab:hardware_specs}
\begin{tabular}{@{}ll@{}}
\toprule
Property & Value \\
\midrule
CPU model & AMD Ryzen Threadripper PRO 7995WX 96-Cores \\
Sockets & 1 \\
Cores per socket & 96 \\
Threads per core & 2 \\
Node memory & 1.0 TiB \\
Threads per process & 1 \\
Requested memory & 80 GiB \\
Python version & 3.12.9 \\
\bottomrule
\end{tabular}
\end{table}